\documentclass[review]{elsarticle}

\usepackage{multirow}
\usepackage[noend]{algpseudocode}
\usepackage{algorithmicx,algorithm}
\usepackage{xcolor}
\usepackage{amsmath,amssymb,amsfonts}
\usepackage{graphicx}
\usepackage{hyperref}
\usepackage{caption}
\usepackage{subfigure}

\journal{Journal of \LaTeX\ Templates}

\begin{document}

\begin{frontmatter}

\title{Rethinking Samples Selection for Contrastive Learning: Mining of Potential Samples}

\author{Hengkui Dong}
\author{Xianzhong Long\corref{mycorrespondingauthor}}
\cortext[mycorrespondingauthor]{Corresponding author}
\ead{lxz@njupt.edu.cn}

\author{Yun Li}

\address{School of Computer Science $\&$ Technology, School of Software \\ Nanjing University of Posts and Telecommunications, Nanjing, China, 210023}

\begin{abstract}
Contrastive learning predicts whether two images belong to the same category by training a model to make their feature representations as close or as far away as possible. In this paper, we rethink how to mine samples in contrastive learning, unlike other methods, our approach is more comprehensive, taking into account both positive and negative samples, and mining potential samples from two aspects: First, for positive samples, we consider both the augmented sample views obtained by data augmentation and the mined sample views through data mining. Then, we weight and combine them using both soft and hard weighting strategies. Second, considering the existence of uninformative negative samples and false negative samples in the negative samples, we analyze the negative samples from the gradient perspective and finally mine negative samples that are neither too hard nor too easy as potential negative samples, i.e., those negative samples that are close to positive samples. The experiments show the obvious advantages of our method compared with some traditional self-supervised methods. Our method achieves 88.57\%, 61.10\%, and 36.69\% top-1 accuracy on CIFAR10, CIFAR100, and TinyImagenet, respectively.
\end{abstract}

\end{frontmatter}


\section{Introduction}
The quality and quantity of the training data are crucial for the model training, but a large amount of labeled data is required at a high cost. Self-supervised learning (SSL) was proposed to address this problem, and nowadays, self-supervised learning is developing rapidly, which can be said to be a promising path for the development of machine learning \cite{new1}.

Self-supervised learning mining useful information from abundant unlabeled data for learning. In the field of computer vision, self-supervised learning, in some cases, even surpasses many supervised methods. Traditional supervised learning methods are based on available labeled data and are trained on specific tasks, which are usually known prior knowledge. Self-supervised learning, on the other hand, can discover more general and intrinsic representations of many tasks. Self-supervised learning has made considerable progress not only in the field of images, but also in corresponding successful applications from natural language processing and now multi-modal, video, audio, and other field \cite{b25,b26,b1,b2,new2,new3},.

Contrastive learning, one of the most successful paradigms of self-supervised learning, trains a model to predict whether two images come from the same class to make their feature representations as close as possible or as far away as possible. Since self-supervised learning has no labeled information, in order to recognize similar inputs, we usually use data augmentation transformation to generate new samples that bring about dissimilarity while maintaining the underlying semantics. We usually consider such input samples to be positive samples, we also want to generate dissimilar samples to prevent the model from collapsing, i.e., negative samples.

Typical self-supervised contrastive learning contains both positive and negative samples. The challenge lies in how to select and mine reasonable and appropriate potential samples for model training. Using augmented samples as positive samples will ignore other positive samples in the dataset and thus lack diversity, but sampling positive samples from mined data will make it difficult to guarantee the authenticity of the samples (i.e., whether or not they belong to the same category as the query samples). The same dilemma exists in negative samples. A negative sample that is too difficult may be a false negative sample, and one that is too easy may be meaningless for model training.

In this paper, based on the above problems, we propose a comprehensive framework for potential sample mining (PSM) based on an asynchronous asymmetric network structure. First, in potential positive samples mining (PPSM), we consider the augmented sample view and the mined sample views, respectively, in order to ensure the authenticity and diversity of the positive samples, and ultimately weight them with hard and soft ways. Second, in potential negative sample mining (PNSM), we consider the mined negative samples should be neither too hard nor too easy and eventually mine the negative samples that are similar to positive samples based on the analysis from the gradient perspective.

All our contributions are listed as follow:
\begin{itemize}
\item We propose a method to mine potential samples in contrastive learning. Our method comprehensively considers both positive and negative samples and shows excellent performance on different datasets.
\item  
By analyzing the purity of positive samples during the training period, we find the defects of the previous methods for mining positive samples, so we consider both mined samples and augmented samples at the same time and  then evaluate the final loss by weighting them.
\item Inspired by the gradient changes of samples during the training, we mine potential negative samples based on how close the negative samples are to the positive sample. In this way, the negative samples can provide more information and is also less likely to be false negative samples.
\end{itemize}

\section{Related work}
 \subsection{Contrastive Learning}
 For each input instance image $x$, its different data augmented variants are treated as positive samples, i.e., augmented input views $(x_1,x_2)$. The other categories of instances are treated as negative samples. The augmented input views are fed into the feature encoder $f$, which are then passed through the projection head $p$ to finally get the projected feature representation $(z_1,z_2)$, $(z_1,z_2)$ are viewed as a pair of positive sample views. 

We denote $\mathcal{P}$ as the set of positive samples, which includes sample features that are related to the query sample $x_i$ and belong to the same instance level; $\mathcal{N}$ is the set of negative samples, which contains the sample features of the other classes. Based on different strategies, the negative samples may come from different places, such as the current batch or memory bank \cite{b3,b4,b5}. The conventional InfoNCE loss function can be defined as follows: 
\begin{equation}
\begin{aligned}
\mathcal{L} _{CL}= 
 -\log\frac{  \exp(s( z_1,z_2)/t )           }    { \exp(s( z_1,z_2)/t )
     +\sum_{i=1}^{\left |\mathcal{N}\right |}\exp(s( z_1,z_i^{-})/t )}   
\label{equation1}
\end{aligned}
\end{equation}
where $s(z_1,z_2)$ denotes the metric function between $z_1$ and $z_2$, usually we use cosine similarity as the similarity metric. $z_i^{-}$ is the negative sample corresponding to the query sample and  $t$ is the temperature parameter for the loss function.

Generally, contrast learning will involve positive and negative samples, and the mining of samples will be presented in two aspects.

\subsection{Samples Mining in Contrastive Learning}
\paragraph{Potential Positive Samples Mining}
Nearest-neighbor sample mining has been widely and maturely used in computer vision. In contrastive learning, NNCLR\cite{b12} mines potential positive samples from the dataset through nearest neighbors in the feature space, arguing that this provides more semantic information than predefined data augmentation. Similarly, MSF\cite{b13} shift sample features are close to the mean of the mined nearest neighbor samples based on BYOL\cite{b6}. What's more, MYOW\cite{b14} employs a cascaded dual projector architecture to predict the mined view through different parts of the network. SNCLR\cite{b15} obtains potential positive samples by mining soft neighbors in the candidate neighbor set and calculates the correlation between predicted samples and soft neighbors by cross-attention module.

\paragraph{Potential Negative Samples Mining}
There are broadly two types of problems in mining negative samples: one is the false negative sample problem and the other is the negative samples do not provide sufficient challenges to model training. For false negative samples, IFND\cite{b16} and FNC\cite{b17} utilize incremental and constructive support views to help false negative elimination and false negative attraction, respectively. DCL\cite{b19}, on the other hand, removes false negative samples with a certain probability by means of debiasing. For hard negative samples, MoChi\cite{b18} generates hard negative samples by Mixup strategy, and HCL\cite{b20} assigns greater mining weights to the harder negative samples by weighting. BCL\cite{b21} is to correct the bias of the original sampling under the Bayesian framework to ensure the reasonableness of mining the samples.

 \subsection{Simplification of Contrastive Learning}
Due to the complexity of sample mining in contrastive learning, there are a few methods that optimize and simplify contrastive learning. BYOL\cite{b6} and SimSiam \cite{b7} both do not use negative samples, but instead prevent the model from collapsing through the use of techniques such as prediction head, asynchronous network, and stopping gradient updating. Barlow Twins\cite{b8} and VICreg\cite{b9} are not as cumbersome as the other methods, they optimize the objective from the perspective of loss function by reducing redundancy through regularization. There are also methods that simplify the contrastive loss in based on temperature parameters and coupling term in the loss function\cite{b10,b11}.

In addition to the above, there are many other methods, such as SwAV\cite{b22}, that utilize clustering prototypes to mine potential samples from a clustering perspective. There are also methods that construct input views to mine different positive samples from a data perspective. Some obtain new input sample views by the Mixup method\cite{b23}, some generate new pairs of positive samples by improving the cropping method\cite{b31}, and some consider both global and local information to construct inputs with multiple views\cite{b32}. Moreover, in \cite{b24},  samples are constructed  from both positive and negative samples at the same time. Similarly, our approach integrates both positive and negative samples and proposes a rational strategy for sample mining respectively.


\begin{figure*}[!t]
\centering  
\subfigure[The purity by top-1 method.]{   
\begin{minipage}{6cm}
\centering    
\includegraphics[scale=0.22]{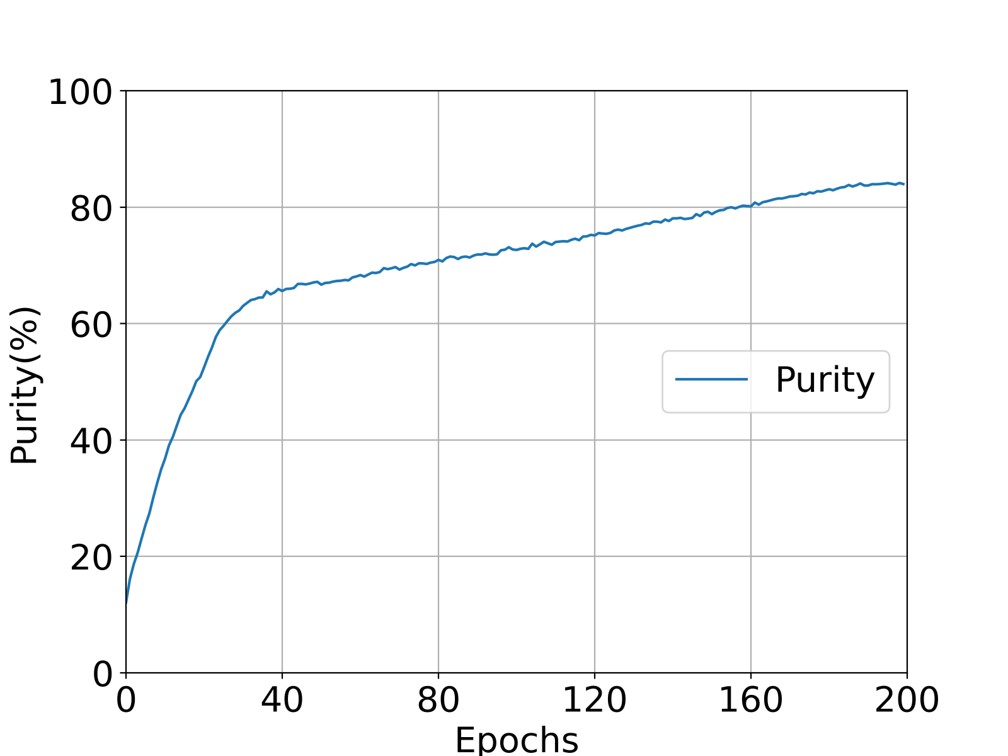}  
\label{fig1a}
\end{minipage}}\subfigure[The purity by top-5 method.]{ 
\begin{minipage}{6cm}
\centering    
\includegraphics[scale=0.22]{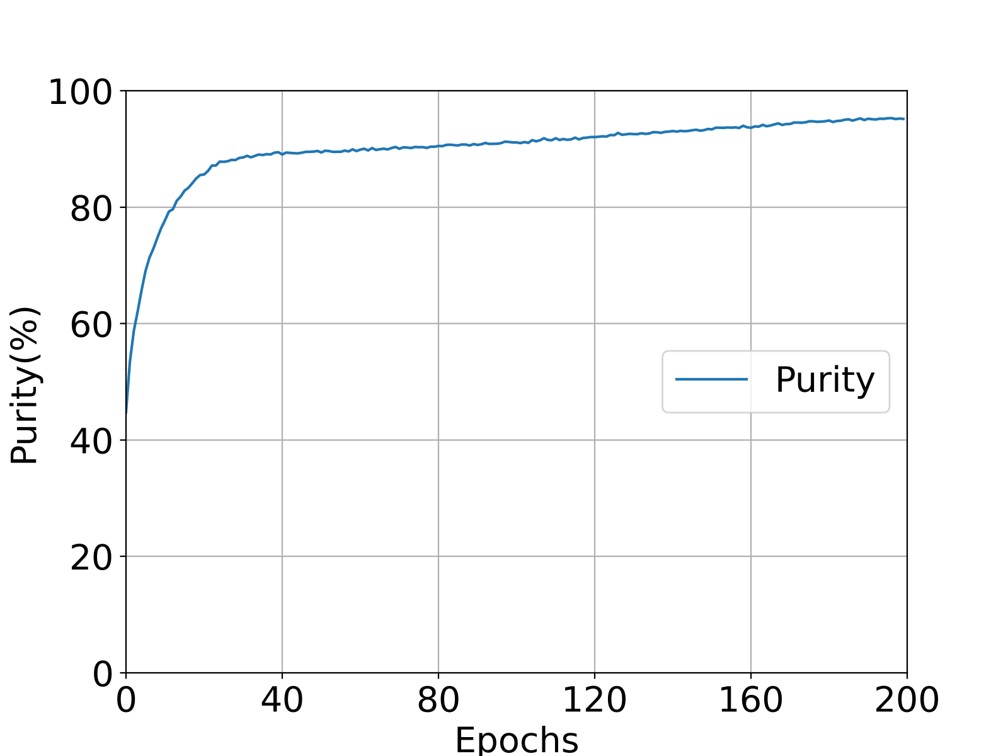}
\label{fig1b}
\end{minipage}
}
\caption{The purity of mined positive samples during model training.}    
\label{fig1}    
\end{figure*}

\section{Rethinking the Effect of Samples Mining in Contrastive Learning}
\label{chap3}
\subsection{The Problem of False Positive Samples in Positive Samples Mining}\label{chap3.1}
Potential positive samples in contrastive learning may come from the augmented sample views or from the mined sample views. NNCLR\cite{b12} mines the sample that are most similar to the query sample as positive sample. This approach is too risky and is likely to result in the mined positive sample being false positive sample.

We define the variable “purity” to measure the risk of positive sample mining. The "purity" of positive samples refers to the proportion of true positive samples mined by the top-$k$ method out of the total number of samples in a batch, i.e., the likelihood of mining true positive samples. As shown in
Figure \ref{fig1a}, the initial purity value of positive sample mined by the top-1 method is quite low. This is attributed to the network's instability during the early stages of training, it is difficult to guarantee the final mining results is a real positive sample, which affects the convergence of the model. With the increase of epochs, the model is progressively trained, leading to a gradual improvement in the purity value. However,  ultimately, the purity cannot be fully guaranteed.

If we increase the number of mined sample views, from mining the most similar to the top-5 most similar, then the probability of mined sample views containing real positive samples is much higher, which alleviates the problem of mining false positive samples to some extent, and also provides richer and more diversified information for the model. SNCLR\cite{b15} is trained by such an idea by utilizing the soft neighbors and cross-attention module to train the network. However, from Figure \ref{fig1b}, we can find that even in the top-5 way of mining, in the early stage of network training, the value of purity is not high. For this reason, our method takes into account both the augmented sample views and the mining views, and then through the mixing of the weights, mining the potential positive samples. On the one hand, the authenticity of the positive samples is guaranteed, and on the other hand, the diversity of the samples is enhanced.

\subsection{The Dilemma of Negative Samples Mining in Contrastive Learning}
\label{chap3.2}
In self-supervised contrastive learning, the challenge of mining negative samples has always been troublesome. There are uninformative, too easy negative samples, and too hard false negative samples. The potential negative samples we need are those that are neither too hard nor too easy.

In the field of computer vision, few studies have analyzed the quality of negative samples from the perspective of gradient. For simplicity of the following analysis, we utilizing the binary cross entropy (BCE) loss instead of the multi-objective classification InfoNCE loss. When the activation function used for the BCE loss is Sigmod function, the gradient of the network model parameters can be written in the following form: 
\begin{equation}
\begin{aligned}
\nabla_{\theta}\mathcal{L} _{BCE}= 
\left\{\begin{matrix}
( (s( z_i,z_k)-1)\nabla_{\theta}s(z_i,z_k)  & if \thinspace  z_k\in  \mathcal{P} 
 \\ 
s( z_i,z_k)\nabla_{\theta}s( z_i,z_k)  &if \thinspace   z_k\in  \mathcal{N}  
\end{matrix}\right.
\label{equation2}
\end{aligned}
\end{equation}
where $z_i$ is the query sample feature representation, $z_k$ is the sample feature used to compute the similarity metric with $z_i$, both of them are normalized and the range of  $s(z_i,z_k)$ is [0, 1]. See \ref{app1} for more details.

When the negative samples are uninformative negative samples, $s(z_i,z_k)$ will be low, i.e., $s(z_i,z_k)\rightarrow 0$, their gradient $\nabla_{\theta}\mathcal{L} _{BCE}$ also tends to zero, so such negative samples have no meaningful role in the training convergence of the model. When the negative sample is false negative sample, the right terms $\nabla_{\theta}s( z_i,z_k)$ are roughly the same, but one is greater than zero on the left, while the other is smaller than zero on the left. Therefore, it will lead to a large variance of the gradient, which will make the parameters of the model unstable, and the model is difficult to be optimized.

\begin{figure*}[!t]
\centering
\includegraphics[scale=0.35]{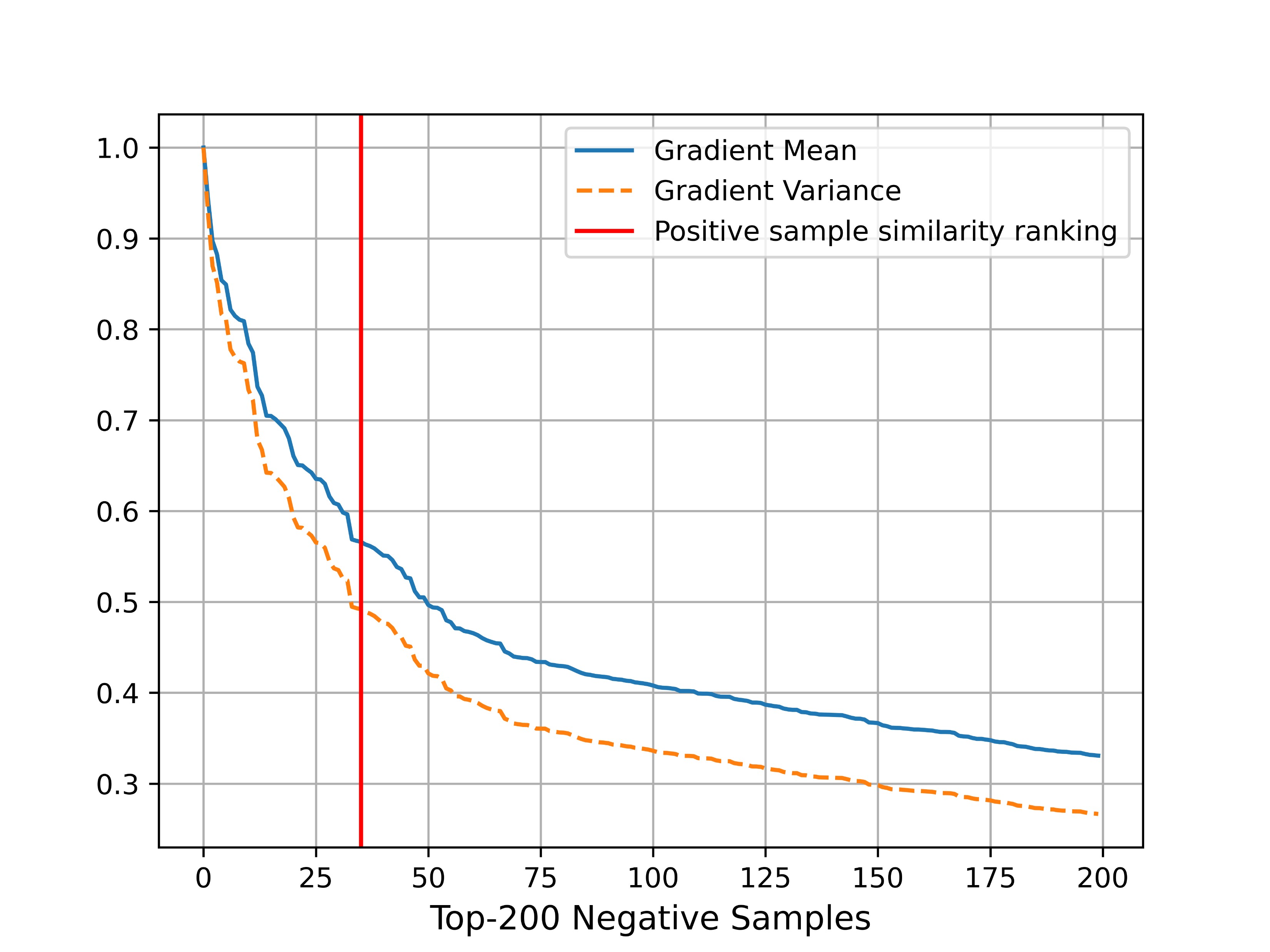}
\caption{The mean and variance change curves of the gradient.}
\label{fig2}
\end{figure*}

We hope to utilize the relationship between positive samples, negative samples, and query samples to help us eliminate unnecessary and uninformative negative samples, so as to mine valuable negative samples.  As shown in Figure \ref{fig2}, we enumerate the top-200 negative samples based on the cosine similarity between the negative samples and the query sample, while marking the similarity ranking of the positive samples (red line). To better show the trend of similarity, we normalized the mean and variance of gradient by its maximum value. 

As can be seen from Figure \ref{fig2}, when the ranking decreases, the mean and variance of the gradient decreases. The top-25 negative samples have significantly larger gradient mean and variance than the other negative samples. The larger gradient mean indicates that these negative samples are harder and more important for model training, while the larger gradient variance, the higher the probability of false negative samples and the more unstable the model. Surprised to find that the negative samples ranked around the positive samples can produce larger gradient and smaller gradient variance, which indicates that these negative samples are more likely to be real negative samples, and at the same time contain more complex semantic information. Based on the above findings, we mine potential negative samples that are similar to positive samples based on their metric similarity.

\section{Method}
\begin{figure*}[!t]
\centering
\includegraphics[scale=0.45]{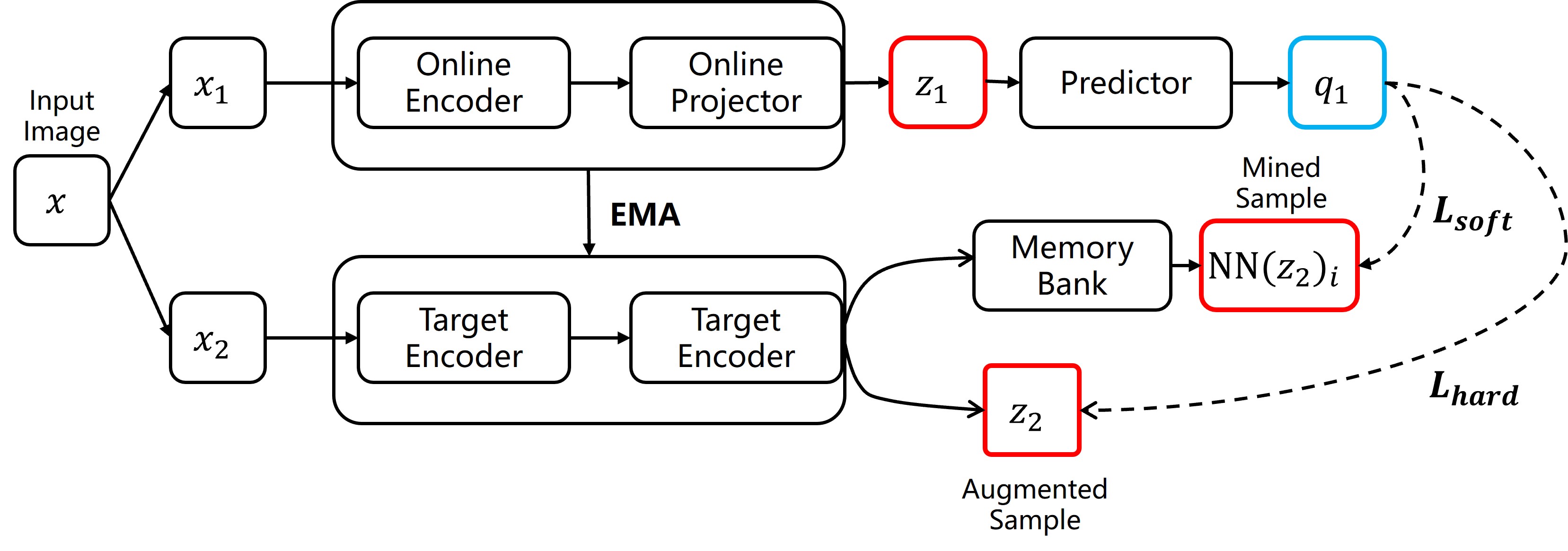}
\caption{The architecture of PSM. Given an input image $x$, the augmented input view ($x_1, x_2$) is obtained sequentially, and then pass through the online network and the target network respectively, which is updated by the exponential moving average (EMA).  The augmented sample view $z_1$ through the prediction head to get the predicted feature representation $q_1$, and $z_2$  goes through the memory bank to mine potential samples $\rm NN(z_2)_i$ . }
\label{fig3}
\end{figure*}

Figure \ref{fig3} shows an overview of PSM, our model improves on the BYOL\cite{b6}. The augmented input images $(x_1, x_2)$ are transformed by the online network and the target network to obtain the projected features $(z_1, z_2)$. We introduce memory bank to help us mine potential sample views, and additional prediction head further transform $z_1$ into $q_1$.

\subsection{Potential Positive Sample Mining}
\label{chap4.1}
Samples of the same class may be viewed as negative samples in contrastive learning, consequently, we obtain positive samples by mining. For each projection feature $z_1$ of the online branch, the corresponding projection feature $z_2$ of target branch is utilized to mine the expected positive samples in the memory bank, which is updated with FIFO strategy. We utilize the cosine similarity between samples to mine top-$k$ most similar features from memory bank:
\begin{equation}
\begin{aligned}
 {\rm NN}(z_2)=\underset{z_i}{{\arg} \,  {\rm top_k}}\{s(z_2,z_i)\mid \forall i\in \{ 1,\dots   ,\left |M \right |\}   \},{\rm where}  \left | {\rm NN}(z_2)\right |=k    
     \label{equation3}
\end{aligned}
\end{equation}
where ${\rm NN}(z_2)$  denotes the set of $k$ feature samples mined from memory bank that are most similar to $z_2$. $\left |M \right |$ is the size of memory bank.

 \paragraph{Hard Mining of Positive Samples} The augmented sample view $z_2$ is an alternative feature representation of the same class of sample $z_1$. For a determined positive sample $z_2$, we use a hard weighting approach, i.e., the loss weight  $w_0$ is set to 1. We define the positive sample hard mining loss as follows:
 
\begin{footnotesize}
\begin{equation}
\begin{aligned}
\mathcal{L} _{hard}= 
 -w_0\log\frac{  \exp(s( q_1,z_2)/t )           }    {\exp(s( q_1,z_2)/t )
     +\sum_{i=1}^{2(N-1)}\exp(s( q_1,z_2^{i-})/t )}   
     \label{equation4}
\end{aligned}
\end{equation}
\end{footnotesize}we use the prediction sample feature $q_1$ of the query sample and its corresponding augmented sample's projected feature $z_2$ to compute the loss, and $N$ is the size of the batch.

 \paragraph{Soft Mining of Positive Samples} 
 From Section \ref{chap3.1},  we understand that blindly selecting the most similar samples as positive samples will bring bad effects in the early stage of model training. When increasing the number of positive samples to be mined, the probability of mining the real positive samples will be greatly increased, and  simultaneously, more positive samples will bring  diversity to the model. To take multiple positive samples into account at the same time, instead of  treating the mined samples as positive samples directly, we assign weight $w_i$ to each mined positive sample when calculating the contrastive loss, $w_i$ is calculated as follows:
\begin{equation}
\begin{aligned}
w_i=\frac{\exp(s(z_1,{\rm NN}(z_2)_i))}{\sum_{j=1}^{k}\exp(s(z_1,{\rm NN}(z_2)_j)))}  
     \label{equation5}
\end{aligned}
\end{equation}

we measure the importance of each mined sample based on the similarity between ${\rm NN}(z_2)_i$ and $z_1$ to the total similarity, $\sum_{i=1}^{k}w_i =1$. The positive sample soft mining loss is as follows:
\begin{footnotesize}
\begin{equation}
\begin{aligned}
\mathcal{L} _{soft}= 
 -\sum_{i=0}^{k}w_i\log \frac{  \exp(s( q_1,{\rm NN}(z_2)_i)/t )           }    { \sum_{m=0}^{k}\exp(s( q_1,{\rm NN}(z_2)_m)/t )
     +\sum_{l=1}^{N-1}\sum_{j=0}^{k}\exp(s( q_1,{\rm NN}(z_2^{l-})_j)/t )}   
     \label{equation6}
\end{aligned}
\end{equation}
\end{footnotesize}where $\rm NN(z_2)_0$ is the augmented sample view $z_2$ itself. Negative samples in $\mathcal{L} _{soft}$ also come from samples mined from memory bank.

\subsection{Potential Negative Sample Mining}
Based on the analysis in Section \ref{chap3.2},  we consider the negative samples that rank close to the positive sample to be the potential negative samples according to the similarity metric, which are neither too hard to be false negative samples nor too easy to be low-quality negative samples. We want such negative samples to be mined during the training of contrastive learning. To this end, we propose a probability of mining negative samples: 
\begin{equation}
\begin{aligned}
p_i(q_1;z_i^{-})=\exp(-a(s(q_1,z_i^{-})-s(q_1,z_2))^{2}),\forall z_i^{-} \in \mathcal{N} 
     \label{equation8}
\end{aligned}
\end{equation}
$s(q_1,z_2)$ denotes the similarity score between the query sample and the positive sample, and $s(q_1,z_i^{-})$ denotes the similarity score between the query sample and the negative sample, we utilize the difference between them as a criterion for mining. $a$ is a hyper-parameter that controls the density of the probability distribution. As the negative samples are closer to the positive sample, the probability of being mined is higher.

\begin{figure*}[!t]
\centering

\includegraphics[scale=0.5]{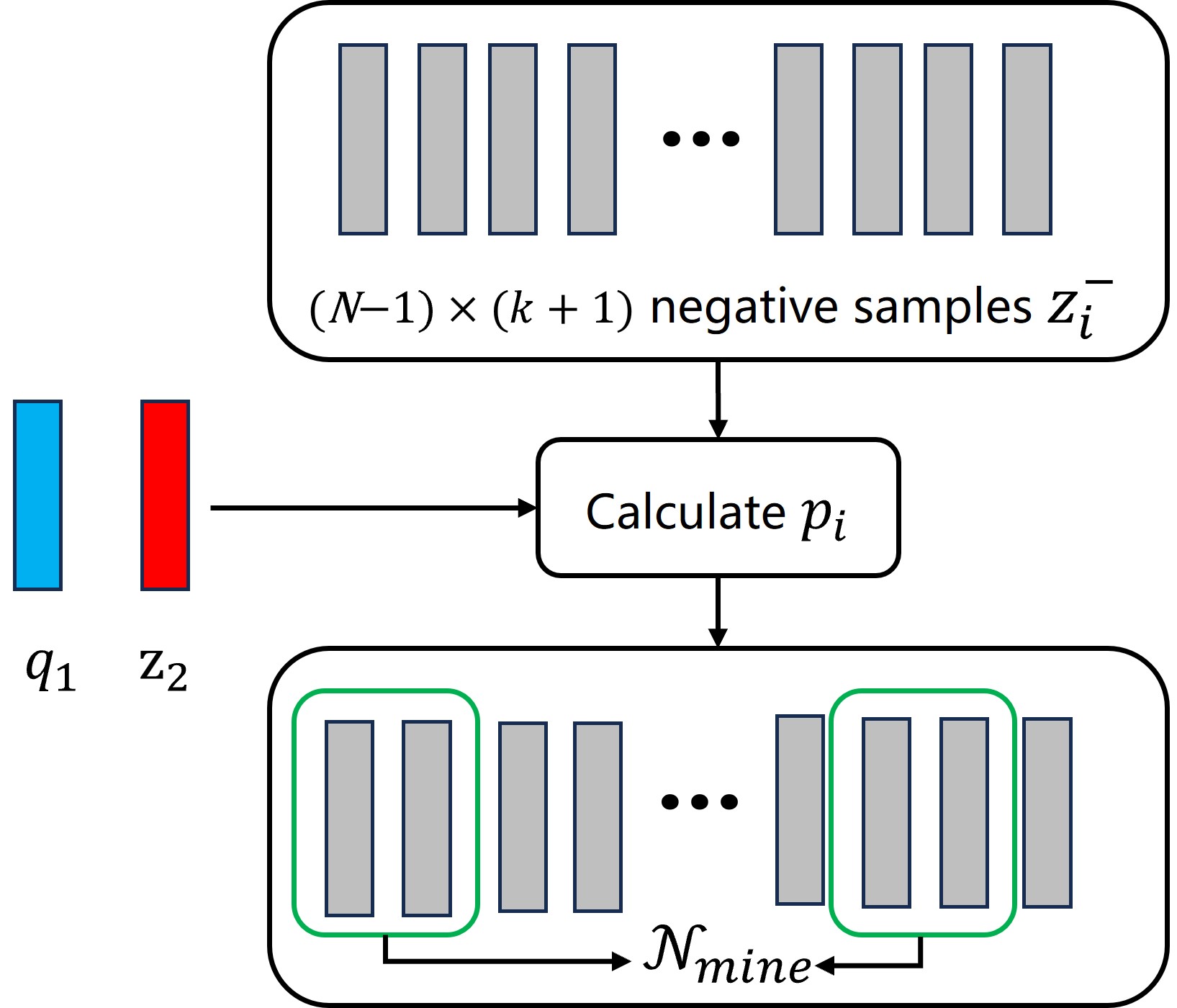}
\caption{Illustration of potential negative samples mining (PNSM).}
\label{fig4}
\end{figure*}

After that, Bernoulli sampling is performed for each negative sample, and finally the specific negative samples corresponding to the query sample $q_1$ are mined: 
\begin{equation}
\begin{aligned}
\mathcal{N}_{mine}(q_1)=\{ \mathcal{B}(z_i^{-};p_i(q_1;z_i^{-}))\mid z_i^{-}\in \mathcal{N}    \}
     \label{equation9}
\end{aligned}
\end{equation}
$\mathcal{B}(z,p)$ denotes the Bernoulli experiment that accepts $z$ with probability $p$. Figure \ref{fig4} illustrates the flow of PNSM, potential negative samples are mined for each query sample in the current batch in turn, and eventually the potential negative samples corresponding to each query sample are mined and the useless negative samples are discarded. For each query sample, the number of its corresponding negative samples is $ (N-1)\times(k+1) $.

\subsection{Overall Loss}
Taking the PPSM and PNSM together, the term in the denominator of  
 Equation (\ref{equation4}) and (\ref{equation6}) with respect to negative samples can be transformed as follows: 
\begin{equation}
\begin{aligned}
\sum_{i=1}^{2(N-1)}\exp(s( q_1,z_2^{i-})/t ) \xrightarrow{mine}
\sum_{i=1}^{\left |\mathcal{N}_{mine}^{hard}\right |}\exp(s( q_1,z_i^{-}/t )
     \label{equation10}
\end{aligned}
\end{equation}
 
\begin{equation}
\begin{aligned}
\sum_{l=1}^{N-1}\sum_{j=0}^{k}\exp(s( q_1,{\rm NN}(z_2^{l-})_j)/t ) \xrightarrow{mine}
\sum_{i=1}^{\left |\mathcal{N}_{mine}^{soft}\right |}\exp(s( q_1,z_i^{-}/t )
     \label{equation10}
\end{aligned}
\end{equation}where $\mathcal{N}_{mine}^{hard}$ and $\mathcal{N}_{mine}^{soft}$ denote the final negative sample set after PNSM mining, respectively. The overall loss of our model is as follows:
\begin{equation}
\begin{aligned}
\mathcal{L} _{PSM}= 
\mathcal{L} _{soft}+\lambda \mathcal{L} _{hard}
     \label{equation7}
\end{aligned}
\end{equation}
where $\lambda$ is the hyper-parameter that controls the importance of $\mathcal{L} _{hard}$ in the overall loss $\mathcal{L} _{PSM}$, we set $\lambda =1 $.

\section{Experiments}
\subsection{Experimental Details}
\paragraph{Datasets}
We conduct experiments on three commonly used benchmark datasets CIFAR10\cite{b27}, CIFAR100\cite{b27} and TinyImageNet\cite{b28}. Both CIFAR10 and CIFAR100 contain 50,000 training images and 10,000 test images. In TinyImaget, we utilize 90,000 of these images as the training set and the remaining 10,000 as the testing set. The training set is utilized in the training process for pre-training without labels and linear fine-tuning with labels. The testing set is used in the final evaluation to assess the performance of the model.

\paragraph{Network Architecture}
We use restnet18\cite{b29} as the backbone encoder of the model, and the input image passes through the encoder to end up with a 512-d feature representation. Both the projection head MLP and the prediction head MLP are composed of two fully connected layers with sizes [512, 512, 128] and [128, 512, 128], respectively. Each fully connected layer is followed by batch-normalization \cite{b30} and ReLU activation function. The last layer has no ReLU activation function.

\paragraph{Experimental Settings and Implementation details}
In order to ensure the fairness of the experiments, the generic parameters of the models involved in the experiments are kept consistent. For the image augmentation transform, we follow the settings in \cite{b3,b5}. We use the SDG optimizer to optimize the network for updates. The models are pre-trained for 200 epochs, with a warm-up period of 20 epochs at the beginning, where the learning rate ranges from 0.0001 to 0.1. Subsequently, the learning rate is decayed by cosine annealing. The batch size is 256 and the temperature parameter $t$ is 0.5. The weight decay is set to 0.001 for the CIFAR10 and CIFAR100, and to prevent overfitting, it is set to 0.0001 for TinyImageNet.

In PSM model, the memory bank size is set to 16834, and the target network is updated using 0.99 momentum. For additional introduced hyperparameters, the top-$k$ parameter is set to 5, the parameter a = 0.5, and the loss weight $\lambda $ = 1.

The experiments of all the methods in this paper are conducted on an Nvidia GTX 3090 GPU.

\begin{table}[]
\caption{Linear evaluation results.}
\label{table1}
\begin{center}
\begin{tabular}{clcclcclcc}
\hline
\textbf{Method}  &  & \multicolumn{2}{c}{\textbf{CIFAR10}} &  & \multicolumn{2}{c}{\textbf{CIFAR100}} &  & \multicolumn{2}{c}{\textbf{TinyImageNet}} \\ \cline{3-4} \cline{6-7} \cline{9-10} 
        &  & Top-1        & Top-5        &  & Top-1         & Top-5        &  & Top-1           & Top-5          \\ \hline
SimCLR \cite{b3}  &  & 82.35        & 99.32        &  & 52.05         & 82.12        &  & 32.23           & 60.42          \\
MoCo v2 \cite{b5} &  & 81.45        & 99.19        &  & 47.71         & 78.64        &  & 33.46           & 61.10          \\
BYOL  \cite{b6}  &  & 86.15        & 99.51        &  & 55.36         & 84.23        &  & 32.60           & 60.62          \\
VICReg \cite{b9} &  & 78.82        & 98.91        &  & 44.53         & 74.34        &  & 27.66           & 52.25          \\
MoChi \cite{b18}  &  & 82.49        & 99.26        &  & 51.49         & 81.36        &  & 33.62           & 61.46          \\
DCL  \cite{b19}   &  & 84.56        & 99.38        &  & 53.75         & 83.03        &  & 33.52           & 60.94          \\
HCL   \cite{b20} &  & 85.91        & 99.53        &  & 58.71         & 85.46        &  & 35.60           & 63.05          \\
BCL   \cite{b21}  &  & 85.16        & 99.38        &  & 57.31         & 85.07        &  & 32.33           & 59.11          \\
NNCLR  \cite{b12} &  & 82.25        & 99.27        &  & 52.03         & 81.79        &  & 34.34           & 62.09          \\
SNCLR \cite{b15}  &  & 87.80        & 99.48        &  & 57.93         & 85.37        &  & 35.68           & 63.80          \\
PSM     &  & \textbf{88.57}        & \textbf{99.59}        &  & \textbf{61.10}         & \textbf{87.29}        &  & \textbf{36.69}           & \textbf{64.29}          \\ \hline
\end{tabular}
\end{center}
\end{table}

\subsection{Comparison to State-of-The-Art Methods}
For pre-trained models, we only keep the encoder and then add a fully connected layer for linear classification. Thereafter we fine-tune the linear classification head for 100 epochs.

We use the top-1 accuracy and top-5 accuracy obtained from the linear evaluation as the criteria for model evaluation. We compare the proposed method PSM with some state-of-the-art and relevant methods, and the comparison results are shown in Table \ref{table1}. On three different datasets, our method PSM outperforms the other methods and achieves the best results with 3.17 $\%$ performance improvement in the top-1 accuracy of CIFAR100. This demonstrates the effectiveness of our proposed potential sample mining method. The mining of both positive and negative samples provides more potential samples for the model and brings performance improvement to the model.

\begin{table}[]
\caption{Evaluation results related to potential positive samples mining.}
\label{table2}
\begin{center}
\begin{tabular}{lllllllll}
\hline
Method    & \multicolumn{4}{c}{CIFAR10}                           & \multicolumn{4}{c}{CIFAR100}                          \\ \cline{2-9} 
          & \multicolumn{2}{c}{Top-1} & \multicolumn{2}{c}{Top-5} & \multicolumn{2}{c}{Top-1} & \multicolumn{2}{c}{Top-5} \\ \hline
PSM       & \multicolumn{2}{c}{88.57} & \multicolumn{2}{c}{99.59} & \multicolumn{2}{c}{61.10} & \multicolumn{2}{c}{87.29} \\
w/o $\mathcal{L} _{hard}$ & \multicolumn{2}{l}{88.52 \textcolor{red}{(-0.05)}} & \multicolumn{2}{l}{99.56 \textcolor{red}{(-0.03)}} & \multicolumn{2}{l}{60.30 \textcolor{red}{(-0.80)}} & \multicolumn{2}{l}{86.96 \textcolor{red}{(-0.33)}  } \\

w/o $\mathcal{L} _{soft}$ & \multicolumn{2}{l}{87.30 \textcolor{red}{(-1.27)}} & \multicolumn{2}{l}{99.52 \textcolor{red}{(-0.07)}} & \multicolumn{2}{l}{59.63 \textcolor{red}{(-1.47)}} & \multicolumn{2}{l}{86.09 \textcolor{red}{(-1.20)} } \\ 

\hline
PPSM & \multicolumn{2}{l}{88.20 \textcolor{red}{(-0.37)}} & \multicolumn{2}{l}{99.65 \textcolor{green}{(+0.06)}} & \multicolumn{2}{l}{59.32 \textcolor{red}{(-1.72)}} & \multicolumn{2}{l}{86.52 \textcolor{red}{(-0.77)} } \\ 

\hline
\end{tabular}
\end{center}
\end{table}

\subsection{Evaluation}
\subsubsection{Evaluation of Potential Positive Samples Mining}
\paragraph{The effects of $\mathcal{L} _{soft}$ and $\mathcal{L} _{hard}$}
The overall loss of our method PSM consists of two parts: $\mathcal{L} _{soft}$ and  $\mathcal{L} _{hard}$. In order to verify the effect of each part of the loss on the whole, we conduct experiments on CIFAR10 and CIFAR100 with $\mathcal{L} _{soft}$ and  $\mathcal{L} _{hard}$ removed, respectively. We can see from Table \ref{table2} that the accuracy of the model decreases without $\mathcal{L} _{soft}$ and  $\mathcal{L} _{hard}$ , and the decrease is even more pronounced without  $\mathcal{L} _{soft}$. This suggests that both components have some enhancement effect on the model.

\begin{figure*}[!t]
\centering  
\subfigure[CIFAR10]{   
\begin{minipage}{6cm}
\centering    
\includegraphics[scale=0.33]{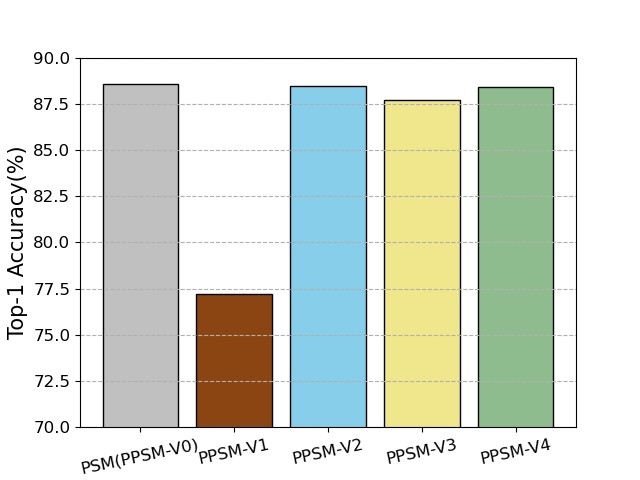}  
\label{fig5a}
\end{minipage}}\subfigure[CIFAR100]{ 
\begin{minipage}{6cm}
\centering    
\includegraphics[scale=0.33]{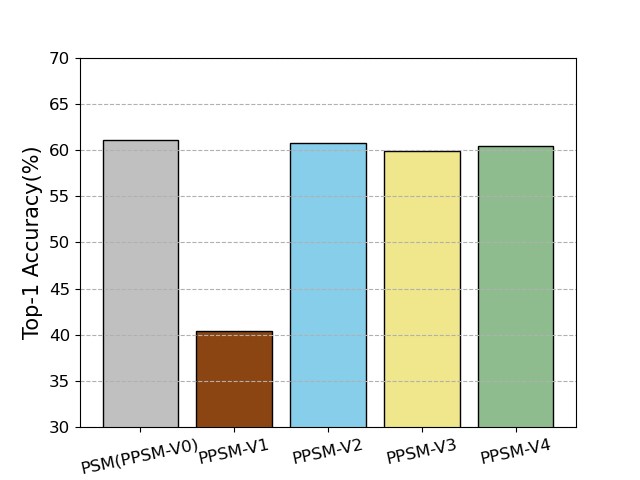}
\label{fig5b}
\end{minipage}
}
\caption{Comparison of different weighting strategies on CIFAR10 and CIFAR100.}    
\label{fig5}    
\end{figure*}

\paragraph{Weighting Strategies for Mining Potential Positive Samples}
To further study the advantage of our PPSM, we compare PSM with 4 variants of soft mining weighted approach and draw the following observations based on the results in Figure \ref{fig5}. \textbf{PSM(PPSM-V0)}: The version used in this paper, i.e. the weights 
calculated by Equation \ref{equation5}, is used as a benchmark; 
\textbf{PPSM-V1}: Setting the threshold = $1/k $ on top of PPSM-V0, ultimately removing mined samples with weights less than $ 1/k$ and retaining mined samples with thresholds greater than $1/k$. The accuracy decreases severely;
\textbf{PPSM-V2}: Further assign further weights to the filtered $k^{'}$ samples on the basis of PPSM-V1 and set the weight of each mined sample to $1/k^{'}$. The accuracy slightly decline;
\textbf{PPSM-V3}: Further weights are assigned to the filtered $k^{'}$ samples based on  PPSM-V1, with the weight of each sample set to 1, which reduces the accuracy.
\textbf{PPSM-V4}: Set the weight of all mined samples to 1. The accuracy decreases slightly.

Summarizing the above observations, our weighting approach can better capture the relationship between mined samples and can better assign diverse weights to different samples.

\begin{table}[ht]
\caption{Ablation experiments with different methods of applying PNSM.}
\vspace{+1.0em}
\label{table3}
\begin{tabular}{llllllll}
\hline
                                &    & \multicolumn{2}{c}{SimCLR \cite{b3}} & \multicolumn{2}{c}{MoCo v2 \cite{b5}} & \multicolumn{2}{c}{HCL \cite{b20}} \\ \cline{3-8} 
  \multirow{-2}{*}{\textbf{Dataset}}                                               &  \multicolumn{1}{c}{\multirow{-2}{*}{\textbf{Method}}}        & Top-1        & Top-5       & Top-1        & Top-5        & Top-1                              & Top-5                             \\ \hline
\multicolumn{1}{c}{}                          & Baseline & 82.35        & 99.32       & 81.45        & 99.19        & 85.91                              & \textbf{99.53 }                            \\
\multicolumn{1}{c}{\multirow{-2}{*}{CIFAR10}} & w/PNSM    & \textbf{83.92}        & \textbf{99.50}       & \textbf{82.52}        & \textbf{99.30}        & \textbf{85.97}                              & 99.52                             \\ \hline
                                              & Baseline & 52.02        & 82.12       & 47.71        & 78.64        & 58.71                              & 85.46                             \\
\multirow{-2}{*}{CIFAR100}                    & w/PNSM    & \textbf{53.82 }       & \textbf{82.93}       & \textbf{50.55 }       &\textbf{80.70}        & \textbf{59.56}                              & \textbf{86.00   }                          \\ \hline
\end{tabular}
\end{table}
\subsubsection{Evaluation of Potential Negative Mining}
In order to verify the effectiveness of potential negative sample mining, we remove the negative sample mining operation on the PSM to obtain the PPSM in Table \ref{table2}, and find that the top-1 accuracies are all decreased, which indicates the effectiveness of the PNSM.

PNSM can also be viewed as a plug-and-play method.  To further validate its performance, we conduct experiments on potential negative sample mining for some methods that require negative samples. As shown in Table \ref{table3}, when PNSM is applied to these methods, their accuracy is roughly improved to some extent.

\subsection{Hyper-Parameter Ablations   }
Our approach introduces three additional hyper-parameters, which are subjected to ablation experiments in order to explore the impact of these parameters on the model performance.
\begin{figure*}[ht]
\centering
\includegraphics[scale=0.35]{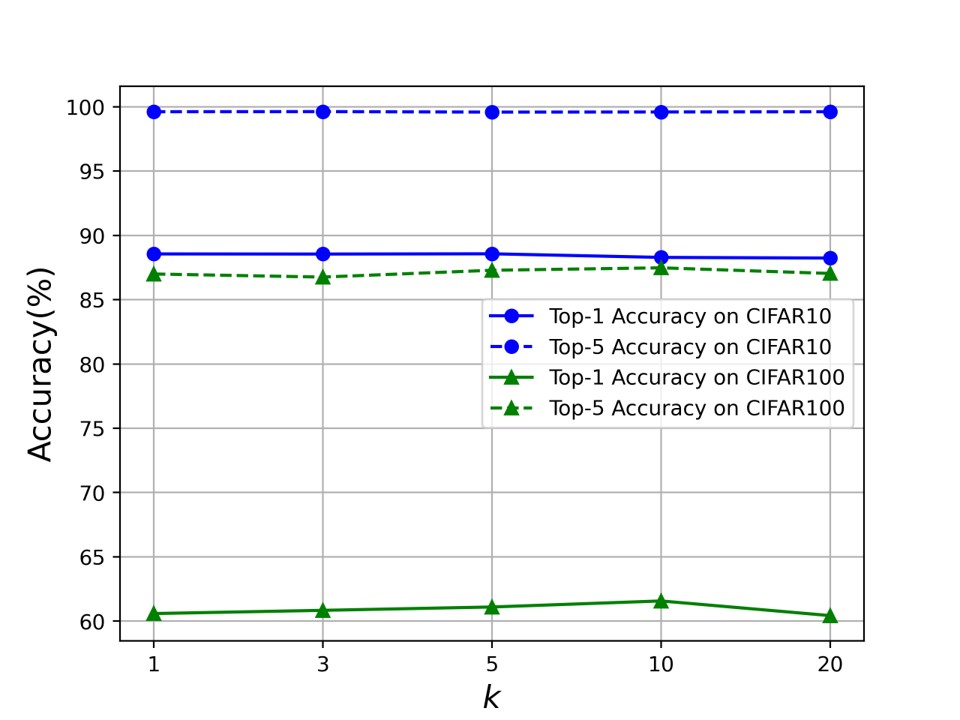}
\caption{Parameter $k$ ablation experiments on CIFAR10 and CIFAR100. }
\label{fig6}
\end{figure*}

\begin{figure*}[!t]
\centering
\includegraphics[scale=0.35]{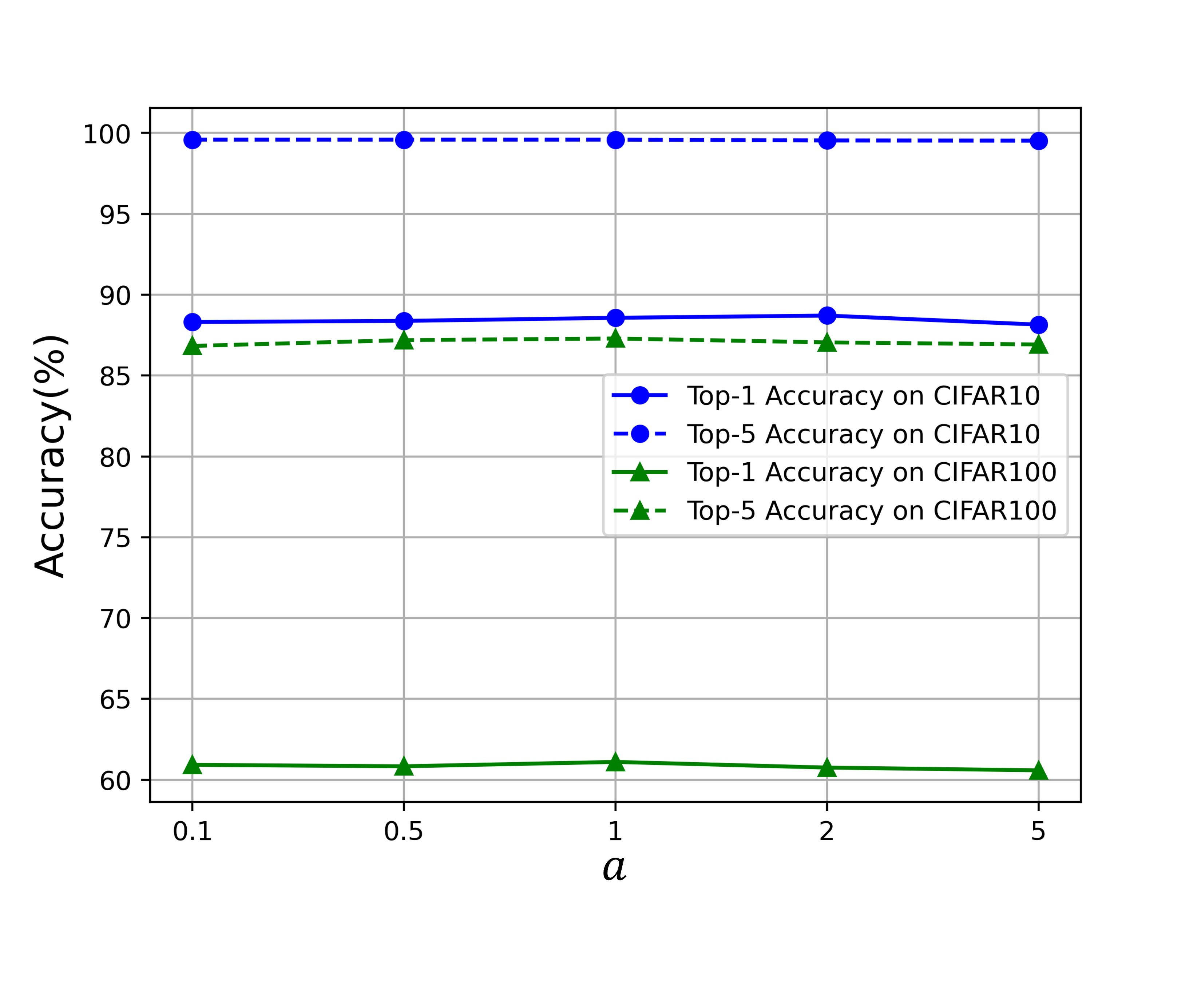}
\vspace{-2.5em}
\caption{Parameter $ a$ ablation experiments on CIFAR10 and CIFAR100. }
\label{fig7}
\end{figure*}

\paragraph{The Impact of Hyper-Parameter $k$}
To explore the effect of mining the number of potential positive samples on the model performance, we set different values of $k$ for the ablation experiments. As shown in Figure \ref{fig6}, on CIFAR100, the top-1 accuracy will increase when k from 1 to 10, but accuracy decreases when k = 20, which we speculate is because mining too many samples may introduce false positive samples.

\begin{figure}[!t]
\centering
\includegraphics[scale=0.35]{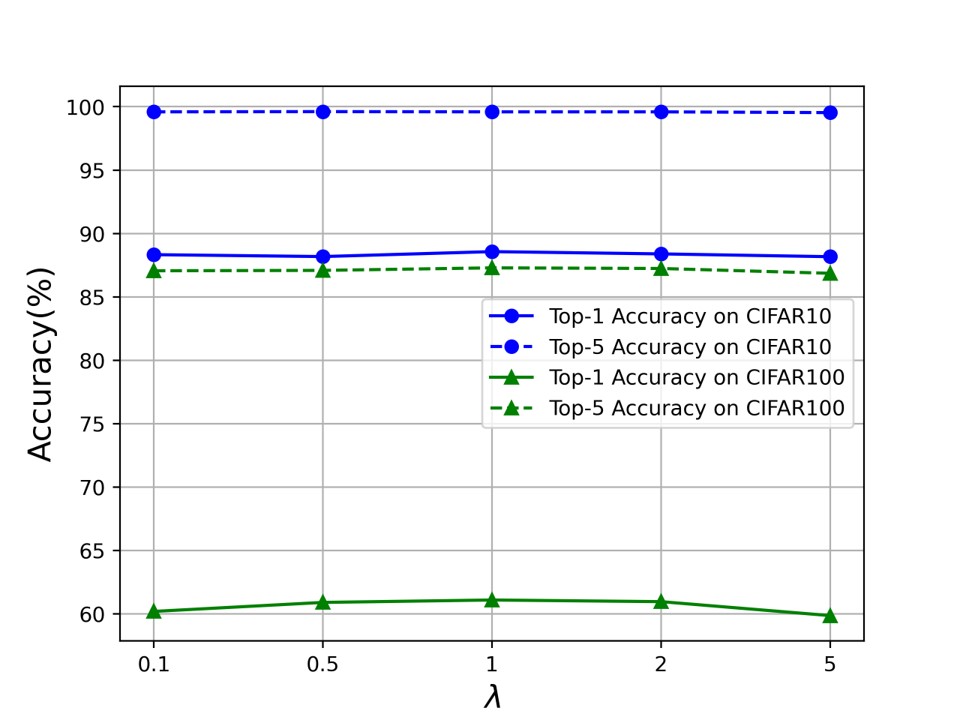}
\caption{Parameter $\lambda$ ablation experiments on CIFAR10 and CIFAR100. }
\label{fig8}
\end{figure}

\paragraph{The Impact of Hyper-Parameters $a$}
Figure \ref{fig7} illustrates the experiment on the CIFAR datasets for different values of a. It can be seen that when a is  appropriately controlled within a certain range, the model is not very sensitive to it.

\paragraph{The Impact of Hyper-Parameters $\lambda$}
Figure \ref{fig8} shows that the best performance is achieved when $\lambda$ = 1, indicating that both $\mathcal{L} _{soft}$ and  $\mathcal{L} _{hard}$ play an important role in the optimization of the model, i.e., both augmented and mined samples need to be considered.

From the three different ablation experiments, in general our model is not very sensitive to hyper-parameters.  Our model has a relatively stable performance for a certain range of parameter variations, which reduces the dependence on specific parameters and is a more reliable and controllable model.

\section{Conclusion}
In this paper, we propose a comprehensive contrastive  learning framework for potential samples mining (PSM). Due to the dilemma of mining positive and negative samples, we rethink how to mine potential samples more rationally in contrastive learning. Specifically, based on the purity of mined positive samples and the change in gradient of negative samples, for positive samples, we propose to utilize both augmented and mined positive samples with a weighted mixture of hard and soft approaches; for negative samples, we mine potential negative samples based on the degree of similarity between negative samples and positive samples. Experiments on all three datasets demonstrate the significant superiority of our approach.

\section*{Acknowledgment}

This work was supported by Postgraduate Research \& Practice Innovation Program of Jiangsu Province KYCX23$\_$1065 and the National Natural Science Foundation of China under Grant No. 61906098.

\bibliography{mybibfile}

\begin{thebibliography}{10}
\expandafter\ifx\csname url\endcsname\relax
  \def\url#1{\texttt{#1}}\fi
\expandafter\ifx\csname urlprefix\endcsname\relax\def\urlprefix{URL }\fi
\expandafter\ifx\csname href\endcsname\relax
  \def\href#1#2{#2} \def\path#1{#1}\fi

\bibitem{new1}
J.~Gui, T.~Chen, Q.~Cao, Z.~Sun, H.~Luo, D.~Tao, A survey of self-supervised
  learning from multiple perspectives: Algorithms, theory, applications and
  future trends, 2023.
\newblock \href {http://arxiv.org/abs/2301.05712} {\path{arXiv:2301.05712}}.

\bibitem{b25}
J.~Devlin, M.~Chang, K.~Lee, K.~Toutanova, {BERT:} pre-training of deep
  bidirectional transformers for language understanding, in: The North American
  Chapter of the Association for Computational Linguistics, 2019, pp.
  4171--4186.

\bibitem{b26}
A.~Radford, J.~W. Kim, C.~Hallacy, A.~Ramesh, G.~Goh, S.~Agarwal, G.~Sastry,
  A.~Askell, P.~Mishkin, J.~Clark, et~al., Learning transferable visual models
  from natural language supervision, in: International Conference on Machine
  Learning, 2021, pp. 8748--8763.

\bibitem{b1}
S.~Liu, A.~Mallol-Ragolta, E.~Parada-Cabaleiro, K.~Qian, X.~Jing, A.~Kathan,
  B.~Hu, B.~W. Schuller, Audio self-supervised learning: A survey, Patterns
  3~(12) (2022) 100616.

\bibitem{b2}
M.~C. Schiappa, Y.~S. Rawat, M.~Shah, Self-supervised learning for videos: A
  survey, ACM Computing Surveys 55~(12) (2022) 1 -- 37.

\bibitem{new2}
C.~Zeng, W.~Wang, A.~Nguyen, Y.~Yue, Self-supervised learning for point cloud
  data: A survey, Expert Systems with Applications (2023) 121354.

\bibitem{new3}
B.~VanBerlo, J.~Hoey, A.~Wong, A survey of the impact of self-supervised
  pretraining for diagnostic tasks with radiological images, 2023.
\newblock \href {http://arxiv.org/abs/2309.02555} {\path{arXiv:2309.02555}}.

\bibitem{b3}
T.~Chen, S.~Kornblith, M.~Norouzi, G.~Hinton, A simple framework for
  contrastive learning of visual representations, in: International Conference
  on Machine Learning, 2020, pp. 1597--1607.

\bibitem{b4}
K.~He, H.~Fan, Y.~Wu, S.~Xie, R.~Girshick, Momentum contrast for unsupervised
  visual representation learning, in: {IEEE/CVF} Conference on Computer Vision
  and Pattern Recognition, 2020, pp. 9729--9738.

\bibitem{b5}
X.~Chen, H.~Fan, R.~Girshick, K.~He, Improved baselines with momentum
  contrastive learning, 2020.
\newblock \href {http://arxiv.org/abs/2003.04297} {\path{arXiv:2003.04297}}.

\bibitem{b12}
D.~Dwibedi, Y.~Aytar, J.~Tompson, P.~Sermanet, A.~Zisserman, With a little help
  from my friends: Nearest-neighbor contrastive learning of visual
  representations, in: {IEEE/CVF} International Conference on Computer Vision,
  2021, pp. 9588--9597.

\bibitem{b13}
S.~A. Koohpayegani, A.~Tejankar, H.~Pirsiavash, Mean shift for self-supervised
  learning, in: {IEEE/CVF} International Conference on Computer Vision, 2021,
  pp. 10326--10335.

\bibitem{b6}
J.-B. Grill, F.~Strub, F.~Altch{\'e}, C.~Tallec, P.~Richemond, E.~Buchatskaya,
  C.~Doersch, B.~Avila~Pires, Z.~Guo, M.~Gheshlaghi~Azar, et~al., Bootstrap
  your own latent-a new approach to self-supervised learning, in: Advances in
  Neural Information Processing Systems, 2020, pp. 21271--21284.

\bibitem{b14}
M.~Azabou, M.~G. Azar, R.~Liu, C.-H. Lin, E.~C. Johnson, K.~Bhaskaran-Nair,
  M.~Dabagia, B.~Avila-Pires, L.~Kitchell, K.~B. Hengen, et~al., Mine your own
  view: Self-supervised learning through across-sample prediction, 2021.
\newblock \href {http://arxiv.org/abs/2102.10106} {\path{arXiv:2102.10106}}.

\bibitem{b15}
G.~Chongjian, J.~Wang, Z.~Tong, S.~Chen, Y.~Song, P.~Luo, Soft neighbors are
  positive supporters in contrastive visual representation learning, in:
  International Conference on Learning Representations, 2023.

\bibitem{b16}
T.-S. Chen, W.-C. Hung, H.-Y. Tseng, S.-Y. Chien, M.-H. Yang, Incremental false
  negative detection for contrastive learning, in: International Conference on
  Learning Representations, 2022.

\bibitem{b17}
T.~Huynh, S.~Kornblith, M.~R. Walter, M.~Maire, M.~Khademi, Boosting
  contrastive self-supervised learning with false negative cancellation, in:
  IEEE/CVF Winter Conference on Applications of Computer Vision, 2022, pp.
  2785--2795.

\bibitem{b19}
C.-Y. Chuang, J.~Robinson, Y.-C. Lin, A.~Torralba, S.~Jegelka, Debiased
  contrastive learning, in: Advances in Neural Information Processing Systems,
  2020, pp. 8765--8775.

\bibitem{b18}
Y.~Kalantidis, M.~B. Sariyildiz, N.~Pion, P.~Weinzaepfel, D.~Larlus, Hard
  negative mixing for contrastive learning, in: Advances in Neural Information
  Processing Systems, 2020, pp. 21798--21809.

\bibitem{b20}
J.~Robinson, C.-Y. Chuang, S.~Sra, S.~Jegelka, Contrastive learning with hard
  negative samples, in: International Conference on Learning Representations,
  2021, pp. 1--29.

\bibitem{b21}
B.~Liu, B.~Wang, Bayesian self-supervised contrastive learning, 2023.
\newblock \href {http://arxiv.org/abs/2301.11673} {\path{arXiv:2301.11673}}.

\bibitem{b7}
X.~Chen, K.~He, Exploring simple siamese representation learning, in:
  {IEEE/CVF} Conference on Computer Vision and Pattern Recognition, 2021, pp.
  15750--15758.

\bibitem{b8}
J.~Zbontar, L.~Jing, I.~Misra, Y.~LeCun, S.~Deny, Barlow twins: Self-supervised
  learning via redundancy reduction, in: International Conference on Machine
  Learning, 2021, pp. 12310--12320.

\bibitem{b9}
A.~Bardes, J.~Ponce, Y.~LeCun, {VICR}eg: Variance-invariance-covariance
  regularization for self-supervised learning, in: International Conference on
  Learning Representations, 2022.

\bibitem{b10}
C.~Zhang, K.~Zhang, T.~X. Pham, A.~Niu, Z.~Qiao, C.~D. Yoo, I.~S. Kweon, Dual
  temperature helps contrastive learning without many negative samples: Towards
  understanding and simplifying moco, in: IEEE/CVF Conference on Computer
  Vision and Pattern Recognition, 2022, pp. 14441--14450.

\bibitem{b11}
C.-H. Yeh, C.-Y. Hong, Y.-C. Hsu, T.-L. Liu, Y.~Chen, Y.~LeCun, Decoupled
  contrastive learning, in: European Conference on Computer Vision, Springer,
  2022, pp. 668--684.

\bibitem{b22}
M.~Caron, I.~Misra, J.~Mairal, P.~Goyal, P.~Bojanowski, A.~Joulin, Unsupervised
  learning of visual features by contrasting cluster assignments, in: Advances
  in Neural Information Processing Systems, 2020, pp. 9912--9924.

\bibitem{b23}
S.~Kim, G.~Lee, S.~Bae, S.-Y. Yun, Mixco: Mix-up contrastive learning for
  visual representation, 2020.
\newblock \href {http://arxiv.org/abs/2010.06300} {\path{arXiv:2010.06300}}.

\bibitem{b31}
X.~Peng, K.~Wang, Z.~Zhu, M.~Wang, Y.~You, Crafting better contrastive views
  for siamese representation learning, in: IEEE/CVF Conference on Computer
  Vision and Pattern Recognition, 2022, pp. 16031--16040.

\bibitem{b32}
T.~Zhang, C.~Qiu, W.~Ke, S.~S{\"u}sstrunk, M.~Salzmann, Leverage your local and
  global representations: A new self-supervised learning strategy, in:
  Proceedings of the IEEE/CVF Conference on Computer Vision and Pattern
  Recognition, 2022, pp. 16580--16589.

\bibitem{b24}
R.~Zhu, B.~Zhao, J.~Liu, Z.~Sun, C.~W. Chen, Improving contrastive learning by
  visualizing feature transformation, in: IEEE/CVF International Conference on
  Computer Vision, 2021, pp. 10306--10315.

\bibitem{b27}
A.~Krizhevsky, G.~Hinton, et~al., Learning multiple layers of features from
  tiny images, 2009, pp. 1--60.

\bibitem{b28}
Y.~Le, X.~Yang, Tiny imagenet visual recognition challenge, CS 231N 7~(7)
  (2015) 3.

\bibitem{b29}
K.~He, X.~Zhang, S.~Ren, J.~Sun, Deep residual learning for image recognition,
  in: {IEEE/CVF} Conference on Computer Vision and Pattern Recognition, 2016,
  pp. 770--778.

\bibitem{b30}
S.~Ioffe, C.~Szegedy, Batch normalization: Accelerating deep network training
  by reducing internal covariate shift, in: International Conference on Machine
  Learning, 2015, pp. 448--456.

\end{thebibliography}

\begin{appendix}
\section{Experimental Details of Section \ref{chap3}}
\label{app2}
\paragraph{Implementation Details on Figure \ref{fig1}} We utilize SNCLR\cite{b15} as the model framework, and in the process of training, we mine samples from memory bank by top-1 and top-5 methods respectively, search for the mined samples that have the same labels as the query samples, and take the mean value for each batch as the result of purity.
\paragraph{Implementation Details on Figure \ref{fig2}}We utilize MoCo V2\cite{b7} to calculate the similarity between query samples and negative samples in memory bank in turn,  and finally enumerate the 200 negative samples with the largest similarity, as well as mark the rank of positive sample similarity among them. For the negative sample gradient mean, first calculate the gradient of negative samples, then according to the Equation \ref{equation2} to get the loss gradient, and finally use the maximum value of the normalization. The gradient variance is calculated in the same way.

\section{Proof of Equation \ref{equation2}  }
\label{app1}
\begin{equation}
\begin{aligned}
\mathcal{L} _{BCE}=-y_{i,k}\log (s(z_i,z_k))-(1-y_{i,k})\log(1-s(z_i,z_k))
\end{aligned}
\end{equation}
where $s(z_i,z_k)$ predicts the probability of $z_k$ as a positive sample of $z_i$  using the metric function $s$, $y_{i,k}$ is the groundtruth label.

\begin{equation}
\begin{aligned}
\mathcal{L} _{BCE}= 
\left\{\begin{matrix}
 -\log (s(z_i,z_k))   & if  \thinspace  y_{i,k}=1
 \\ 
 -\log (1-s(z_i,z_k))  & if  \thinspace   y_{i,k}=0  
\end{matrix}\right.
\end{aligned}
\end{equation}
each gradient with respect to $s(z_i,z_k)$ depends only on the loss given by its binary problem. When activation function is the sigmoid function, the gradient with respect to $s$ can be written as:

\begin{equation}
\begin{aligned}
\frac{\partial \mathcal{L} _{BCE}}{\partial s} = y_{i,k}(s(z_i,z_k)-1)+(1-y_{i,k})s(z_i,z_k)
\end{aligned}
\end{equation}
where $y_{i,k}=1$ means $z_i$ and $z_k$ belong to the same class, i.e., $ z_k\in  \mathcal{P}  $.

\begin{equation}
\begin{aligned}
\frac{\partial \mathcal{L} _{BCE}}{\partial s} = \left\{\begin{matrix}
 s(z_i,z_k)-1  & if  \thinspace  z_k\in  \mathcal{P} \\ 
 s(z_i,z_k)  & if  \thinspace z_k\in  \mathcal{N}
\end{matrix}\right.
\end{aligned}
\end{equation}

\begin{equation}
\begin{aligned}
\nabla_{\theta}\mathcal{L} _{BCE}= \frac{\partial \mathcal{L} _{BCE}}{\partial s}\nabla_{\theta}s(z_i,z_k) =
\left\{\begin{matrix}
( s( z_i,z_k)-1)\nabla_{\theta}s( z_i,z_k)  & if $\thinspace$  z_k\in  \mathcal{P} 
 \\ 
s( z_i,z_k)\nabla_{\theta}s( z_i,z_k)  &if $\thinspace$   z_k\in  \mathcal{N}  
\end{matrix}\right.
\end{aligned}
\end{equation}

\end{appendix}

\end{document}